# Optimal Bangla Keyboard Layout using Association Rule of Data Mining

Md. Hijbul Alam, Abdul Kadar Muhammad Masum, Mohammad Mahadi Hassan and

S M Kamruzzaman

Department of Computer Science & Engineering
International Islamic University Chittagong, Chittagong-4203, Bangladesh
Email: hijbul_bd@yahoo.com, akmmasum@yahoo.com, mahadi_cse@yahoo.com, smk_iiuc@yahoo.com

**Abstract**

*In this paper we present an optimal Bangla Keyboard Layout, which distributes the load equally on both hands so that maximizing the ease and minimizing the effort. Bangla alphabet has a large number of letters, for this it is difficult to type faster using Bangla keyboard. Our proposed keyboard will maximize the speed of operator as they can type with both hands parallel. Here we use the association rule of data mining to distribute the Bangla characters in the keyboard. First, we analyze the frequencies of data consisting of monograph, digraph and trigraph, which are derived from data wire-house, and then used association rule of data mining to distribute the Bangla characters in the layout. Finally, we propose a Bangla Keyboard Layout. Experimental results on several keyboard layout shows the effectiveness of the proposed approach with better performance.*

## 1. INTRODUCTION

The usefulness of computer is increasing rapidly. It is the fastest growing industry in the world. New invention of technologies makes it faster. The billions of instructions can be done within a second nowadays. But the technology of input device like keyboard has not been changed much. The typing speed is very slow in respect of computer's other devices. Its main cause is the complex and insufficient keyboard layout and human limitations [6] . A scientific keyboard layout with equal hands load and maximum hand-switching can reduce this problem. The use of Bangla s increasing day by day in everyday life. Specially in data entry and printing sector. But there is no scientific Bangla keyboard layout at present and very few research works have been done in this field. Data mining popularly referred to as Knowledge Discovery in Databases (KDD), is the automated extraction of patterns representing knowledge implicitly stored in large databases, data warehouses, and other massive information repositories. In this paper we have tried to design a Bangla keyboard layout using the association rule of data mining. We have collected the data from various Bangla documents and we have used the association rule to extract the association of Bangla letters each other. And finally we have designed a Bangla keyboard layout with equal hands load and maximum hand switching.

## 2. MINING ASSOCIATION RULE

Association rule mining finds interesting association or correlation relationships among a large set of data items. The discovery of interesting association relationships among huge amounts of transaction records can help in many decision-making processes. Let us consider the following assumptions for representing the Association rule in terms of mathematical representation, **J** = $\{i_1, i_2, \ldots, i_m\}$ be a set of items. **D** = Set of database transactions where each transaction T is a set of items such that T $\subseteq$ J. Each transaction is associated with an identifier, called TID. **A, B** = Set of items. A transaction T is said to contain A if and only if A $\subseteq$ T. An association rule is an implication of the form A $\Rightarrow$ B, where A $\subset$ J, B $\subset$ J, and A $\cap$ B = $\varnothing$ The rule A $\Rightarrow$ B holds in the transaction set D with support *S*, where *S* is the percentage of transaction in D that contain A $\cup$ B, i.e., Support (A $\Rightarrow$ B) = P (A $\cup$ B). The rule A $\Rightarrow$ B has confidence *C* in the transaction set D if *C* is the percentage of transaction in D containing A that also contain B, i.e., confidence (A $\Rightarrow$ B) = P (B | A) = [support count(A $\cup$ B) / support count(A)]. [1],[3]

## 3. METHODOLOGY:

At first we calculate the monograph, digraph from the frequencies of characters from the Bangla documents. And then we find the association of a character with the determined left hand typed letter sets and right hand typed letter sets according to association rule and distribute it in opposite side for better or maximize hand switching as described in the algorithm at section 4. Finally, we distribute the most frequent word closer to finger for easy and first access. Before stating the algorithm we have to explain some keyword in brief. The meaning of support and confidence is stated earlier in section 2. We introduce four new terms for our purpose. Left support of a letter means the cumulative support of the letter to the set of left sets letter. Left confidence of a letter means the cumulative confidence of the letter to the right sets letter. Similarly right support and confidence carry the meaning as well.

## 4. PROPOSED ALGORITHM

1. Find out the most frequent item set of monograph (letter) in descending order
2. Distribute 1st and 4th letter in the right hand (as 1st letter found twice of second)
3. Distribute 2nd and 3rd letter in the left hand
4. Now take the letter one by one starting from 5th letter and distribute it in left hand or right depend on the following criteria
   a. Find the cumulative support and confidence of this character with all right hand letter distributed till now and labeled it as right support and right confidence
   b. Similarly find left support and left confidence for the same letter
   c. if left support > right support and left confidence > right confidence then distribute the letter in right side otherwise in left side.
5. Repeat the step 4 until all letters are distributed.

## 5. EXPERIMENTAL RESULT

About more than 8 lakh letters are extracted from various types of documents such as prose, text book, novel, religious book etc. From this data we find out monographs, digraphs and then determine support and confidence to mine the association.

Table 1: The first ten monograms of our experiment

| Letter | Frequency | Percentage |
|---|---|---|
| া | 74300 | 9.039875 |
| ে | 45525 | 5.538901 |
| র | 41844 | 5.091044 |
| ি | 37010 | 4.502904 |
| ক | 31214 | 3.797721 |
| ই | 28996 | 3.527863 |
| ব | 28212 | 3.432476 |
| ত | 21451 | 2.609884 |
| প | 18419 | 2.240989 |
| ম | 17202 | 2.092920 |

So according to the 1-3 steps of the algorithm we can initialize the left and right side characters, which are as follows: Right { া, ি }, left { ে, র } Now we have to take the decision in which set the 5th character will lies. The decision is depend on the step 4.

Table 2: The association of ka(ক) with some letters are as follows

| Digraph | Frequency | Support | Confidence |
|---|---|---|---|
| কে | 8316 | 1.011785 | 21.717897 |
| কা | 8000 | 0.973338 | 20.892638 |
| কর | 4134 | 0.502972 | 10.796271 |
| কি | 3094 | 0.376438 | 8.080228 |
| এক | 2062 | 0.250878 | 5.385077 |
| তক | 1231 | 0.149772 | 3.214855 |
| কল | 1153 | 0.140282 | 3.011151 |

The support and confidence of K with left item set { ে, র } is

| কে | 1.011785 | 21.717897 |
|---|---|---|
| কর | 0.502972 | 10.796271 |
| (After addition) | 1.514757 | 32.514168 |

So left support (cumulative) is 1.5154757
left confidence (cumulative) is 32.514168
The support and confidence of K with right item set { া, ি } is

| কা | 0.973338 | 20.892638 |
|---|---|---|
| কি | 0.376438 | 8.080228 |
| (After addition) | 1.349776 | 28.972866 |

So right support (cumulative) is 1.349776
Right confidence is 28.972866. In this example left support is greater then right support and left confidence is greater right confidence. Thus ক will go to right item sets. As a result the members of item sets are increased. The sets Right { া, ি, ক } and left { ে, র } Thus all letter are distribute in left and right side according to the descending order of the frequency of the letter. After executing the program we got the following two sets
Right{
া ি ক স ন য দ শ ী ‌ এ থ উ ণ হ ঃ ড় ক় ঠ ৈ চ় ঐ ঞ ঞ্য ছ ঔ
}
Left
{
ে র ন ব ত প ম হ অ । য জ গ ধ ট ভ ং চ ও ঈ খ ঙ ঘ ‌ ঝ ড ী ঋ ঋ চ়
}
Then we distribute the letter according to the most frequent occurring letter in middle row that is closer to the hand as described in the algorithm at step 4. Our designed Bangla keyboard layout is below.

| । অ প ম হ | য ল স দ শ |
|---|---|
| ত ব ে র ন | ে ি া ক |
| ট ধ য জ গ | ী চ |

with shift key

With ctrl key চ ট ঐ ঙ এ উ ঔ

## 6. COMPARATIVE STUDY

Here we show the comparative study and experimental result with our proposed keyboard layout to Bijoy and proposed lay out3 [2] . The layouts are as follows

**Bijoy keyboard layout:**

with shift key

**Proposed keyboard layout 3 [2]:**

With shift key

Table 3: comparative study with other keyboard

| Name | Hand switching | left hand load | Right hand load | Not determine |
|---|---|---|---|---|
| Proposed Optimal keyboard layout | 410113 | 380058 | 340903 | 133290 |
| Bijoy keyboard layout | 358873 | 475556 | 242526 | 138643 |
| Proposed layout 3 [2] | 358672 | 319946 | 363077 | 173702 |

From the above table we can easily find out that our keyboard layout is optimal. Here we work with total 856725 character (with out space). In Bijoy the load in left is very large than right hand. And hand switching is also smaller than us. Increasing of the data this ratio also increased highly. On the other our keyboard layout is optimal because we design the layout by mining the proper association of letter each other in the various kinds of thoroughly. Not only depend on the frequently occurring monograph, digraph, etc. As for why our keyboard layout shows maximum hand switching than also proposed keyboard layout 3[2]. In the above the table not determining characters are large because here we do not consider 0 to 10, united font etc. From the above data it is clear that our proposed layout divides the load on both hands equally and hand switching is also maximize. Comparing with other keyboard layout we can say that, our proposed layout is optimal.

## 7. CONCLUSION

The use of Bangla keyboard layout is increasing day by day. So it is the time to research on this important field. The Bangla academy also can come forward to finalize such an important matter and would produce an optimal keyboard layout and circular throughout Bangladesh for new generation. Our keyboard layout shows maximum hand switching than any other exiting keyboard and also cost effective because frequently occurred letter is distribute closer to finger. If our keyboard layout is implemented, we hope that people will be benefited.